\documentclass[10pt,twocolumn,letterpaper]{article}

\usepackage{iccv}      

\usepackage{pifont}
\usepackage{arydshln}
\usepackage{makecell}
\usepackage{multirow}

\usepackage{microtype}
\usepackage{tcolorbox}

\definecolor{tab-red}{RGB}{248,206,204}
\definecolor{method-green}{RGB}{213,232,212}

\newcommand{\inlineColorbox}[2]{\begingroup\setlength{\fboxsep}{1pt}\colorbox{#1}{\hspace*{2pt}\vphantom{Ay}#2\hspace*{2pt}}\endgroup}

\newcommand{\eli}[1]{\textcolor{magenta}{[Elisa: #1]}}

\def\methodshort{R2P\xspace}
\def\method{Retrieval and Reasoning for Personalization\xspace}
\def\dataset{Personal Concepts with Visual Ambiguity\xspace}
\def\datasetshort{PerVA\xspace}

\newcommand{\vlm}{VLM\xspace}
\newcommand{\vlms}{VLMs\xspace}
\def\yollava{Yo’LLaVA\xspace}

\newcommand{\vlmtext}{\mathcal{E}_T}
\newcommand{\vlmimage}{\mathcal{E}_V}

\newcommand{\simimg}{{s}^{V,V}} 
\newcommand{\simtxt}{{s}^{V,T}} 
\newcommand{\simatt}{{s}^{V,A}} 
\newcommand{\simscore}{s} 
\newcommand{\netvlm}{\Phi_\mathtt{VLM}}
\newcommand{\conceptset}{\mathcal{C}}
\newcommand{\database}{\mathcal{D}}
\newcommand{\refimage}{I}
\newcommand{\featureimage}{f^V}
\newcommand{\featuretext}{f^T}
\newcommand{\queryimage}{Q}
\newcommand{\prompt}{P}
\newcommand{\conceptname}{c}
\newcommand{\predconcept}{\tilde{c}}
\newcommand{\category}{g}
\newcommand{\desc}{d}
\newcommand{\attribute}{a}
\newcommand{\attributes}{A}

\newcommand{\supmat}{\textit{Supp. Mat. \xspace}}

\definecolor{block-gray}{gray}{0.96}

\definecolor{block-gray}{gray}{0.96}
\newtcolorbox{myquote}{colback=block-gray,
grow to right by=-2mm,grow to left by=-2mm, width=0.95\linewidth,
boxrule=0pt,boxsep=0pt}

\definecolor{iccvblue}{rgb}{0.21,0.49,0.74}
\usepackage[pagebackref,breaklinks,colorlinks,allcolors=iccvblue]{hyperref}

\def\paperID{10177} 
\def\confName{ICCV}
\def\confYear{2025}

\title{Training-Free Personalization via Retrieval and Reasoning on Fingerprints}

\author{Deepayan Das\textsuperscript{$1$} \quad
Davide Talon\textsuperscript{$2$} \quad
Yiming Wang\textsuperscript{$2$} \quad
\\
Massimiliano Mancini\textsuperscript{$1$} \quad
Elisa Ricci\textsuperscript{$1,2$} \\
\small
$^1$University of Trento \quad \quad $^2$Fondazione Bruno Kessler \\
\small
\texttt{\{deepayan.das, massimiliano.mancini, e.ricci\}@unitn.it}\\
\small
\texttt{\{dtalon, ywang\}@fbk.eu}\\
}

\begin{document}
\maketitle
\begin{abstract}
Vision Language Models (VLMs) have lead to major improvements in multimodal reasoning, yet they still struggle to understand user-specific concepts. Existing personalization methods address this limitation but
heavily rely on training procedures, that can be either costly or unpleasant to individual users.
We depart from existing work, and for the first time explore the training-free setting in the context of personalization. We propose a novel method, \method{} (\methodshort{}), leveraging internal knowledge of VLMs. 
First, we leverage VLMs to extract the concept \textit{fingerprint}, i.e., key attributes uniquely defining the concept within its semantic class. When a query arrives, the most similar fingerprints are retrieved and scored via chain of thought reasoning. To reduce the risk of hallucinations, the scores are validated through cross-modal verification at the attribute level:
in case of a discrepancy between the scores, \methodshort{} refines the concept association via
pairwise multimodal matching, where the retrieved fingerprints and their images are
directly compared with the query.
We validate \methodshort{} on two publicly available benchmarks and a newly introduced dataset, \dataset{} (\datasetshort{}), for concept identification highlighting challenges in visual ambiguity. \methodshort{} consistently outperforms state-of-the-art approaches on various downstream tasks across all benchmarks.
Code and data are available at the project page: 
\href{https://deepayan137.github.io/papers/training-free-personalization.html}{\texttt{Training-Free Personalization}}.
\end{abstract}    
\section{Introduction}
\label{sec:intro}
\begin{figure}[t!]
    \centering
    \includegraphics[width=\linewidth]{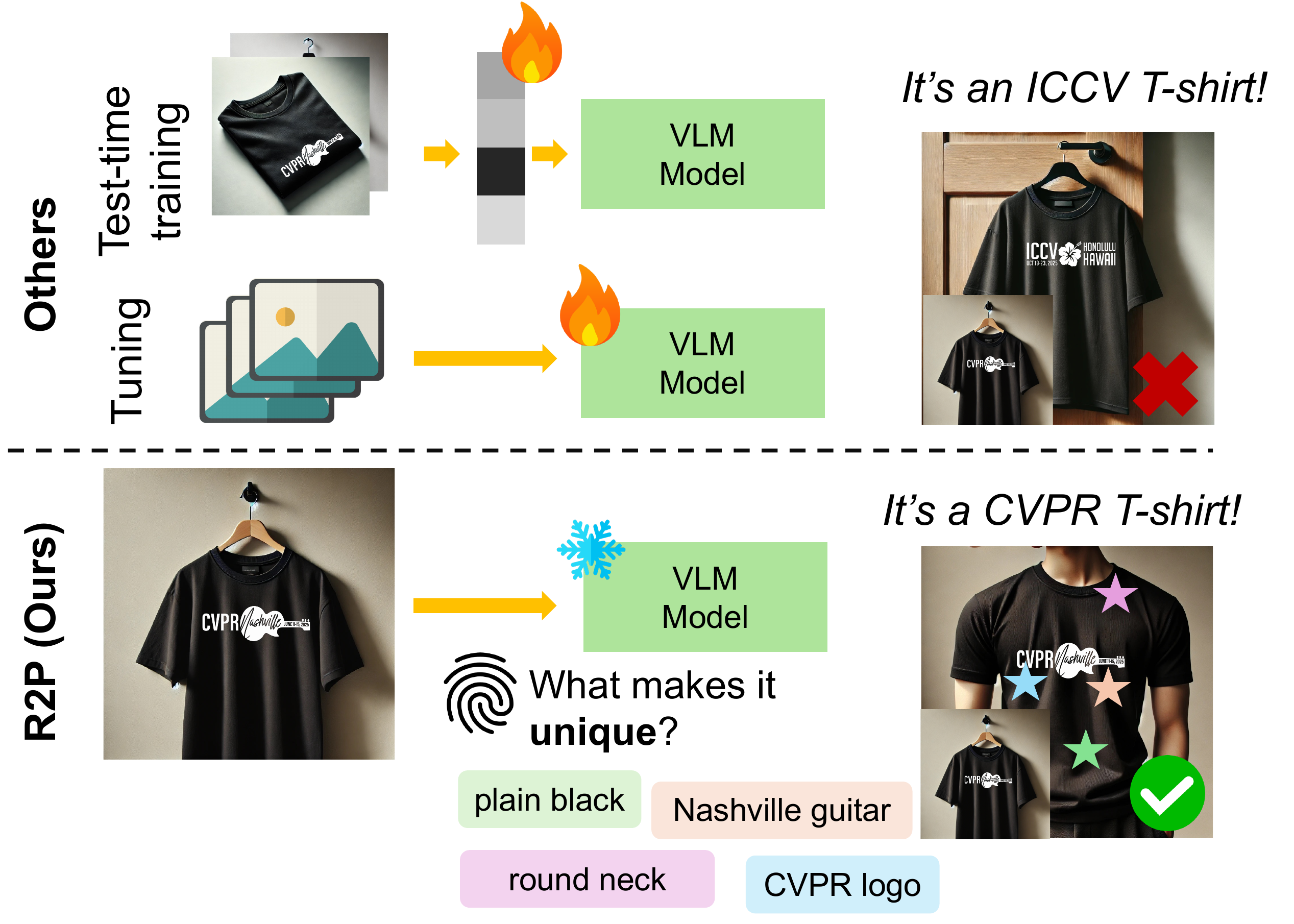}
    \caption{Current \vlms{} personalization methods depend on expensive training procedures. In contrast, we introduce \methodshort{}, a training-free approach that utilizes textual attributes as unique \textit{fingerprints} for identifying personal concepts.
    }
    \label{fig:teaser}
\end{figure}

While "\textit{Where are my keys?}" or "\textit{What is Fuffy doing?}" are questions easy to understand for humans, they are challenging for Vision and Language Models (\vlms)~\citep{liu2024visual, dubey2024llama, abdin2024phi, achiam2023gpt, alayrac2022flamingo} due to their inherent lack of understanding for user-specific concepts. Toward addressing this limitation, 
\vlms personalization aims to add personal concepts to the knowledge of a \vlm~\citep{alaluf2024myvlm, nguyen2024yollava, hao2025remember}. 
Recent works on \vlm personalization mainly rely on training procedures,   
where {multiple reference samples} are used to learn a custom representation for each personal concept in a contrastive manner, against a large number of negative samples from the same class~\citep{nguyen2024yollava, alaluf2024myvlm}. This data collection is not only costly, but 
requires expensive fine-tuning each time a new concept is added. 
Alternatively, other works 
perform expensive large-scale fine-tuning on synthetic personalization data~\citep{hao2025remember, pi2024personalized}, using 
a retrieval-augmented strategy 
to strengthen the user-specific context at inference time~\citep{hao2025remember}. However, they do not eliminate the cost of large-scale pretraining, which can be prohibitive 
when considering modern \vlms. 

One common assumption among these works is that training is needed to enable \vlms to capture 
personal concepts. However, \vlms  have been exposed to virtually any semantic concept through their web-scale training data. The availability of this vast knowledge, raises a natural question: 
\textit{Can we exploit \vlms's internal knowledge to perform personalization without training?} 

In this work, we show for the first time that training-free personalization is feasible, introducing \textit{\method{} (\methodshort{})}, a novel retrieval-reasoning framework to infer personal concepts of interest, using only pre-trained \vlms. 
Specifically, we first construct a database containing rich multimodal information about all user-specific concepts, with reference images and textual descriptions. In particular, for each concept, we leverage the internal world knowledge of a \vlm to enrich the textual description with its distinctive attributes, considering its semantic category and the reference image. Such set of distinctive attributes help in uniquely identifying the concept, serving as a concept \textit{fingerprint}, as shown in~Fig.~\ref{fig:teaser}. Note that the personal database is created once and can be extended flexibly whenever new personal concepts arrive. 
At inference time, given the query image, \methodshort{} first extracts a set of candidate concepts from the personal database, via multimodal retrieval. 
Then, we perform Chain-of-Thought (CoT) reasoning with multimodal prompt, where the \vlm is instructed to focus on \textit{fingerprint attributes} and infer the most-matched personal concept among all retrieved candidates, given their textual information.
As \vlms are notorious for visual hallucination, especially on fine-grained attributes~\cite{zhao2024first,whitehead2022reliable}, we introduce a cross-modal attribute verification module. Specifically, from the CoT reasoning output, we extract the fingerprint attributes that the \vlm leveraged for the concept inference. With pre-trained vision-text aligned encoders~\cite{radford2021learning}, we check the alignment between attributes and the query image, from which we can derive the best-matched personal concept. We consider a positive verification when the inferred concept via attribute-image alignment coincides with the CoT output. In case of a mismatch, the prediction is refined via 
multimodal pairwise reasoning, where the query is compared with each candidate concept (using reference images and textual descriptions). 



We evaluate \methodshort{} on two existing benchmarks for \vlm personalization, MyVLM~\citep{alaluf2024myvlm} and \yollava{}~\citep{nguyen2024yollava}, we further introduce a new personalization dataset, \dataset{} (\datasetshort{}), specifically featuring the challenge of visually and semantically similar concepts.
On all benchmarks, \methodshort{}, while being training-free, consistently outperforms training-based competitors on both recognition, VQA, and captioning tasks, with a significant margin (\eg, +2\% weighted recognition accuracy on \yollava{} Dataset, +8.4\% captioning recall on \datasetshort{}).

\noindent{\textbf{Our contributions}} are summarized as follows:
\begin{itemize}

\item We investigate a novel \textit{training-free setting} for personalization, by leveraging internal knowledge in pre-trained VLMs to eliminate training efforts.

\item We propose a novel method, \textit{\methodshort{}}, which uses a retrieval-reasoning paradigm and textual attributes to uniquely identify personal concepts.


\item We introduce a new benchmark, \textit{\datasetshort{}}, specifically designed to challenge personalization methods by incorporating concepts with high visual and semantic similarity.

\item Experiments demonstrate that \methodshort{} achieves \textit{state-of-the-art} across all benchmarks, successfully recognizing concepts of interest amid visual ambiguity.
\end{itemize}

\section{Related Work}
\label{sec:related}

\paragraph{Personalization with \vlms.} 
Personalization has been addressed on different vision tasks. For instance, in text-to-image generation, target objects are associated with unique identifiers learned from a few provided images, \eg, via textual-inversion~\citep{galimage} or with fine-tuning~\citep{ruiz2023dreambooth, kumari2023multi, shi2024instantbooth}. 
In segmentation, recent approaches leverage correspondences between the reference and test image to construct a confidence map about where the target object is located~\citep{zhang2023personalize, liu2024matcher, tang2024towards, samuel2024wheres} for later prompting with SAM~\citep{kirillov2023segment}. To this end, strategies rely on diffusion models~\citep{tang2024towards, samuel2024wheres}, personalized generative models~\citep{sundaram2024personalized} or cycle consistency~\citep{cohen2022my}. 

In this work we specifically focus on the problem of \vlm personalization where models can recognize and refer to personal concepts of interest.
Earlier approaches, such as MyVLM~\citep{alaluf2024myvlm} and Yo'LLaVa~\citep{nguyen2024yollava}, use an inversion strategy inspired by generative models, where each object is assigned a unique latent representation, either as a concept vector~\citep{alaluf2024myvlm} or token~\citep{nguyen2024yollava}. 
Recent works focused on reducing the time required to personalize VLMs to new concepts by using large-scale tuning and multiple images as inputs to provide in-context information~\citep{hao2025remember, pi2024personalized, pham2024personalized}. However, state-of-the-art approaches either require test-time training or large-scale pretraining, potentially limiting their generalization capabilities. Furthermore, they implicitly assume that personal objects do not change over time. 

In contrast with previous works, we propose leveraging the world knowledge of the VLM to identify unique features of the object of interest, and building on the reasoning capabilities of recent models to achieve personalization in a training-free manner.

\paragraph{Attribute-based inference.} Recognizing objects from the attributes composing them has been a long-standing problem in computer vision~\cite{felzenszwalb2008discriminatively,lampert2009learning}. Notable efforts have been done in the context of zero-shot learning (ZSL), where an attribute classifier trained on a set of classes allows to recognize unseen ones at test time~\cite{lampert2009learning,akata2013label,xian2017zero}. Similarly, researchers studied how attributes change the appearance of an object, both in images~\cite{misra2017red,nagarajan2018attributes,mancini2021open} and in videos~\cite{doughty2020action,doughty2022you,damen2014you}, performing compositional recognition. 

Our work is close in spirit to those aiming to discover useful attributes ~\cite{isola2015discovering,xu2022vgse} or machine generated ones~\cite{pratt2023does,menon2023visual,roth2023waffling,yang2023language} for aiding class discrimination. Moreover, similar to works performing visual classification with contrastive \vlms, we use machine generated descriptions to better recognize a concept. However, differently from these works, we focus on how to describe a personal object to the \vlm, using cross-modal verification to reduce issues on hallucinations~\cite{li2023evaluating,zhou2023analyzing,leng2024mitigating} and compositional reasoning~\cite{yuksekgonuland,hsieh2023sugarcrepe}. 
\section{\method{}}  
\label{sec:method} 
\begin{figure*}[ht!]
    \centering
    \includegraphics[width=\linewidth]{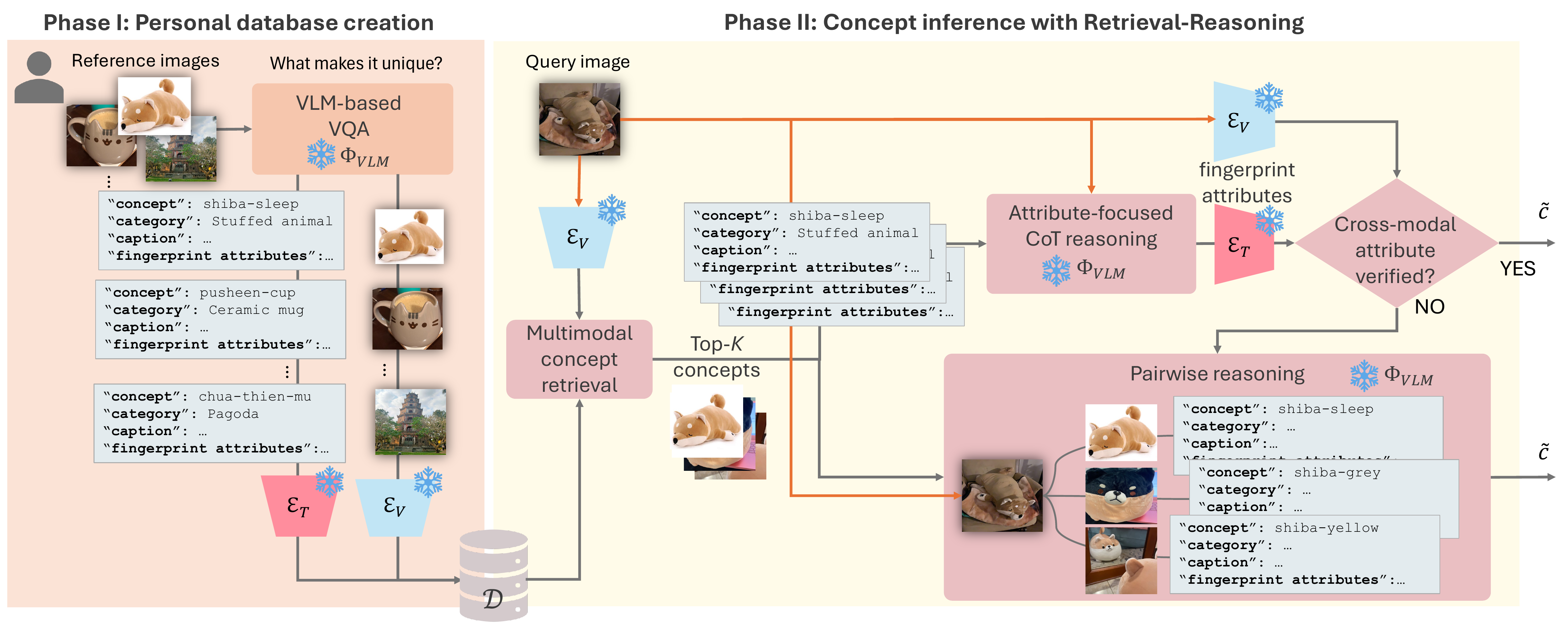}
    \caption{\methodshort{} is the first training-free method to address \vlm personalization, aiming to recognize the personal concept from a query image. \methodshort{} consists of two phases. First, in the the personal database $\database$ creation phase, we leverage the \vlm $\netvlm$ to enrich personal concepts with their distinctive \textit{fingerprint} attributes. Then, in the inference phase, relevant concepts are retrieved from the personal database, and the best matched personal concept $\predconcept$ is obtained with focused reasoning based on images and 
    fingerprint attributes.}
    \label{fig:method}
\end{figure*}
We first introduce the problem of \vlms personalization. Then we describe our training-free method \methodshort{} in details, in terms of personal database creation (Sec.~\ref{sec:method:database}), and personal concept inference with retrieval and reasoning (Sec.~\ref{sec:method:retrieval}).

\noindent\textbf{Problem formulation.}
The goal of personalization is to enable a \vlm to reason about user-specific concepts, \eg, answering queries about them.
Let us denote with $\mathcal{V}$ and $\mathcal{T}$ the visual and textual space, respectively. The \vlm of interest, $\netvlm$, maps images and text inputs to textual outputs, \ie, $\netvlm:\mathcal{V}\times\mathcal{T}\rightarrow\mathcal{T}$.

Following~\cite{hao2025remember}, we assume that the user provides a reference image $\refimage_i \in \mathcal{V}$, a name $\conceptname_i\in\mathcal{T}$ (\eg, ``Sleeping Shiba") for each of the $N$ personal concepts of interest, together with its category $\category_i\in\mathcal{T}$ (\eg, ``Stuffed animal"), from which we can construct a user-specific multimodal database $\database$. 
Given a test-time query image $\queryimage \in \mathcal{V}$ together with a textual prompt $\prompt_q\in \mathcal{T}$ (\eg, ``Is Sleeping Shiba in the image?"), the \vlm $\netvlm$ should provide the answer relevant to the personal concept (\eg, "yes! Sleeping Shiba is resting on the couch"). 
Such personalization process involves two distinct tasks: (i) recognizing which specific personal concept is present in the image and (ii) answering the query. 

Early approaches focus on learning either concept-specific embeddings~\cite{alaluf2024myvlm,nguyen2024yollava}, or a $\netvlm$ that models personal concepts via instruction tuning~\cite{hao2025remember}. Differently, we address the personalization problem in a training-free manner with a novel retrieval-reasoning paradigm. Our proposed method \methodshort{} consists of two phases, as shown in \cref{fig:method}. \textit{Phase I} aims to create the personal database $\database$ via \vlm-based Visual Question Answering (VQA), enriching images with textual information in form of distinctive \textit{fingerprint} attributes. \textit{Phase II} refers to the inference phase for recognizing the personal concept from the personal database. Note that the database is created once and can be flexibly updated with concepts inclusion/deletion/modification. 

\subsection{Personal Database Creation} \label{sec:method:database} 

In the first phase, we aim to construct a multimodal database, $\database$, to store the user-provided personal concepts along with the enriched textual information regarding their distinctive \textit{fingerprint} attributes. 
The \textit{fingerprint} attributes help to uniquely distinguish the personalized concept from other visually similar ones.
The inclusion of such \textit{fingerprint} attributes is a major distinction of our work from prior work~\cite{hao2025remember}, that also involves retrieval at inference time.
%

Specifically, we leverage the same \vlm $\netvlm$ for personalization to extract the \textit{fingerprint} attributes via multimodal prompting.
The reference image $\refimage_{i}$ serves as the visual prompt ${\prompt}^V_D$ and the category $\category_{i}$ is incorporated into the the textual prompt ${\prompt}^T_D$. 
By prompting the \vlm with (${\prompt}^V_D$, ${\prompt}^T_D$), we produce a list of distinctive \textit{fingerprint} attributes $\attributes_{i}$ and a brief caption $\desc_{i}$ aiming to uniquely describe the concept:
\begin{equation}
   \centering
   \{ \attributes_{i},~\desc_{i} \} = \netvlm({\prompt}^V_D,~{\prompt}^T_D).
\end{equation}
Note that the textual prompt ${\prompt}^T_D$ explicitly instructs the \vlm to extract fine-grained attributes $\attributes_{i}$ (\eg, fur color, shirt logo) to help disambiguate similar objects, ensuring that $\attributes_{i}$ contains distinctive attributes, and the description $\desc_{i}$ is discriminative. Please refer to \supmat for the complete multimodal prompt ${\prompt}^V_D,~{\prompt}^T_D$. 

To facilitate retrieval at the inference phase (\textit{Phase II}), we further propose to encode the reference images and descriptions of all personal concepts.
Specifically, for each personal concept, we encode its reference images using an image encoder $\vlmimage:\mathcal{V}\rightarrow \mathcal{Z}$, obtaining the visual embedding $\featureimage_{i} = \vlmimage(\refimage_{i})$, where $\mathcal{Z}$ indicates the latent embedding space. Similarly, we encode the description $\desc_{i}$ using the text encoder $\vlmtext:\mathcal{T}\rightarrow \mathcal{Z}$, obtaining the textual embedding $\featuretext_{i} = \vlmtext(\desc_{i})$. 

At the end of this phase, we have the enriched multimodal personal database $\database={\left\{\refimage_i, \conceptname_{i},~\category_i, \desc_i, \attributes_i, \featureimage_i,\featuretext_i \right\}}^N_{i=1}$, where we included as new elements the extracted visual ($\featureimage_i$) and textual ($\featuretext_i$) features, the enriched descriptions ($\desc_i$), and the set of distinctive fingerprint attributes ($\attributes_i$).
In the following, we describe how we use the database $\database$ during inference. 


\subsection{Concept inference with Retrieval-Reasoning}
\label{sec:method:retrieval}
\paragraph{Multimodal concept retrieval.} At inference phase, given a query image $\queryimage\in\mathcal{V}$, we first retrieve its $K$ most relevant concepts that are visually and textually similar from the database $\database$. 

Specifically, we compute the image embedding of the query image $\featureimage_q$ using the image encoder $\vlmimage$, \ie, $\featureimage_q=\vlmimage(\queryimage)$. Then we calculate the cosine similarity between $\featureimage_q$ and each visual embedding $\featureimage_i \in \database$ and textual embedding of stored personal concepts $\featuretext_i \in \database$:
\begin{align}
    \simimg_{q,i} &= \langle \featureimage_q, \featureimage_i\rangle, \forall \featureimage_i \in \database\\ \nonumber
    \simtxt_{q,i} &= \langle \featureimage_q, \featuretext_i\rangle, \forall \featuretext_i \in \database,
\end{align}
where $\langle x, y\rangle=\frac{x^\mathbf{t} y}{||x|| \cdot||y||}$ is the cosine similarity between two vectors $x$ and $y$.

The final similarity score $\simscore^{q,i}$ accounts for both textual and visual similarities as:
\begin{equation}
    \simscore_{q,i} = \frac{1}{2}(\simimg_{q,i}+ \simtxt_{q,i}), 
    \forall i \in \database.
\end{equation} 

Finally, we select the top-$K$ candidate personal concepts ${\conceptset}^K$ with the highest score $\simscore_{q,i}$.
By incorporating both visual and textual cues, this multimodal retrieval strategy is effective, particularly in cases where visual similarity alone may be misleading (see results of our ablation study in \cref{sec:experiments}).   

\paragraph{Attribute-focused CoT reasoning.} Once we obtain the set of candidate personal concepts $\conceptset^K$, we further leverage the reasoning capabilities of the same \vlm $\netvlm$, to select the concept that best matches the query image, focusing on the \textit{fingerprint} attributes. 
In this initial reasoning step, we prompt the \vlm in a multimodal fashion to first focus on the \textit{fingerprint} attributes that are shared among the query image and then on each candidate concept within $\conceptset^K$. The output is the 
best matched candidate concept $\predconcept$ based on the common attributes. 
Specifically, the query image $\queryimage$ serves as the visual prompt $\prompt_{R}^{V}$ to the \vlm.
The textual prompt $\prompt_{R}^{T}$ includes the textual information of \textit{all the candidate concepts}, followed by the instruction on attribute-based CoT reasoning. Please refer to \supmat for the complete multimodal prompt $\prompt_{R}^{V}, \prompt_{R}^{T}$. 

By feeding both $\prompt_{R}^{V}$ and $\prompt_{R}^{T}$ to the \vlm, we obtain the output concept $\predconcept$, as well as a list of \textit{fingerprint} attributes $\attributes_{q,i}$ shared among the query image and each personal concept $\conceptname_{i}$ within $\conceptset^K$, as expressed below:
\begin{equation}
    \left\{\attributes_{q,i}, \forall i \in\conceptset^k\right\},~\predconcept = \netvlm({\prompt}^V_R,~{\prompt}^T_R).
\end{equation}

\paragraph{Cross-modal attribute verification.}
In the previous step, the \vlm mostly relies on the textual information of the candidate concepts for reasoning. 
While this is computationally efficient, \vlm may struggle with fine-grained disambiguation due to model hallucinations and the lack of visual cues~\cite{zhao2024first,whitehead2022reliable}. 
For instance, some attributes that are leveraged for reasoning might not be present in the query image, making the model prediction less reliable.

Thus, we further validate the predicted concept $\predconcept$ based on the vision-language alignment at attribute level. 
Specifically, for each candidate concept $\conceptname_i \in \conceptset^k$, we encode each \textit{fingerprint} attribute $\attribute_j \in \attributes_{q,i}$ using the text encoder $\vlmtext$, and compute an attribute-based cross-modal embedding similarity score with the image embedding of the query image $\featureimage_q$ accounting all $\attributes_{q,i}$, \ie,
\begin{equation}
    \centering
    \simatt_{q,i} = \frac{1}{|\attributes_{q,i}|}\sum_{\attribute_j \in \attributes_{q,i}}\langle \featureimage_q, \featuretext_{a,j}\rangle,
\end{equation}
where $\featuretext_{a,j} = \vlmtext(\attribute_j)$. 

We can then obtain the concept with the highest attribute-based cross-modal similarity as:
\begin{equation}
    \centering
    \predconcept_a = \arg\max_{\conceptname\in \conceptset^k} \,\,\,\simtxt_{q,i}. 
\end{equation}
If $\predconcept=\predconcept_a$, the global and attribute-based predictions match: thus, the presence of the concept in the image is verified. On the other hand, if $\predconcept\neq \predconcept_a$, the verification fails:
this triggers the more exhaustive but accurate pairwise reasoning, as detailed below.

\paragraph{Pairwise reasoning.} 
Following the verification step, we disambiguate uncertain cases by leveraging $\netvlm$, checking if each concept in $\conceptset^K$ is present in the query image by relying on multimodal information of retrieved personal concepts. 
We frame the recognition problem as a binary classification task where for each candidate concept $\conceptname_i\in\conceptset^K$, we predict how likely $\conceptname_i$ is present in the query image $\queryimage$.


We provide multimodal prompts to the \vlm for each candidate concept. Specifically, the visual prompt $\prompt_{P}^{V}$ includes both the query image $\queryimage$ and the reference image $\refimage_i$. The textual prompt $\prompt_{P}^{T}$ contains all textual information stored in the personal database $\database$ regarding the candidate concept, \ie, $\left\{ \conceptname_{i},~\category_i, \desc_i, \attributes_i \right\}$.
The textual prompt $\prompt_{P}^{T}$ also instructs the model to answer whether the query image matches the currently compared concept in the format of \texttt{Yes/No} and evaluate the probability of the object to be present as:
\begin{equation}
p_i = \frac{\lambda^{\texttt{Yes}}_i}{\lambda^{\texttt{Yes}}_i + \lambda^{\texttt{No}}_i}
\end{equation}
where $\lambda^\texttt{Yes}_i$ and $\lambda^\texttt{No}_i$ denote the logit that the output layer of the \vlm assigns respectively to the \texttt{Yes} and \texttt{No} tokens when answering for the concept $\conceptname_i \in \conceptset^K$. 

Hence, the final predicted concept $\predconcept$ is refined as the candidate concept maximizing the probability $p_i$:
\begin{equation}
    \centering
    \predconcept = \arg\max_{\conceptname\in \conceptset^K} \,\,\,p_i. 
\end{equation}

With the recognized concept $\predconcept$, the \vlm can then generate personalized captions or answer a personalized query, depending on the user prompt. 
Note that when the user explicitly mentions the concept name in his textual prompt, the corresponding image and text description can be retrieved from the $\database$ without going through retrieval-reasoning steps in the proposed \methodshort{}. 


\begin{figure}[t!]
    \centering
    \includegraphics[width=1\linewidth]{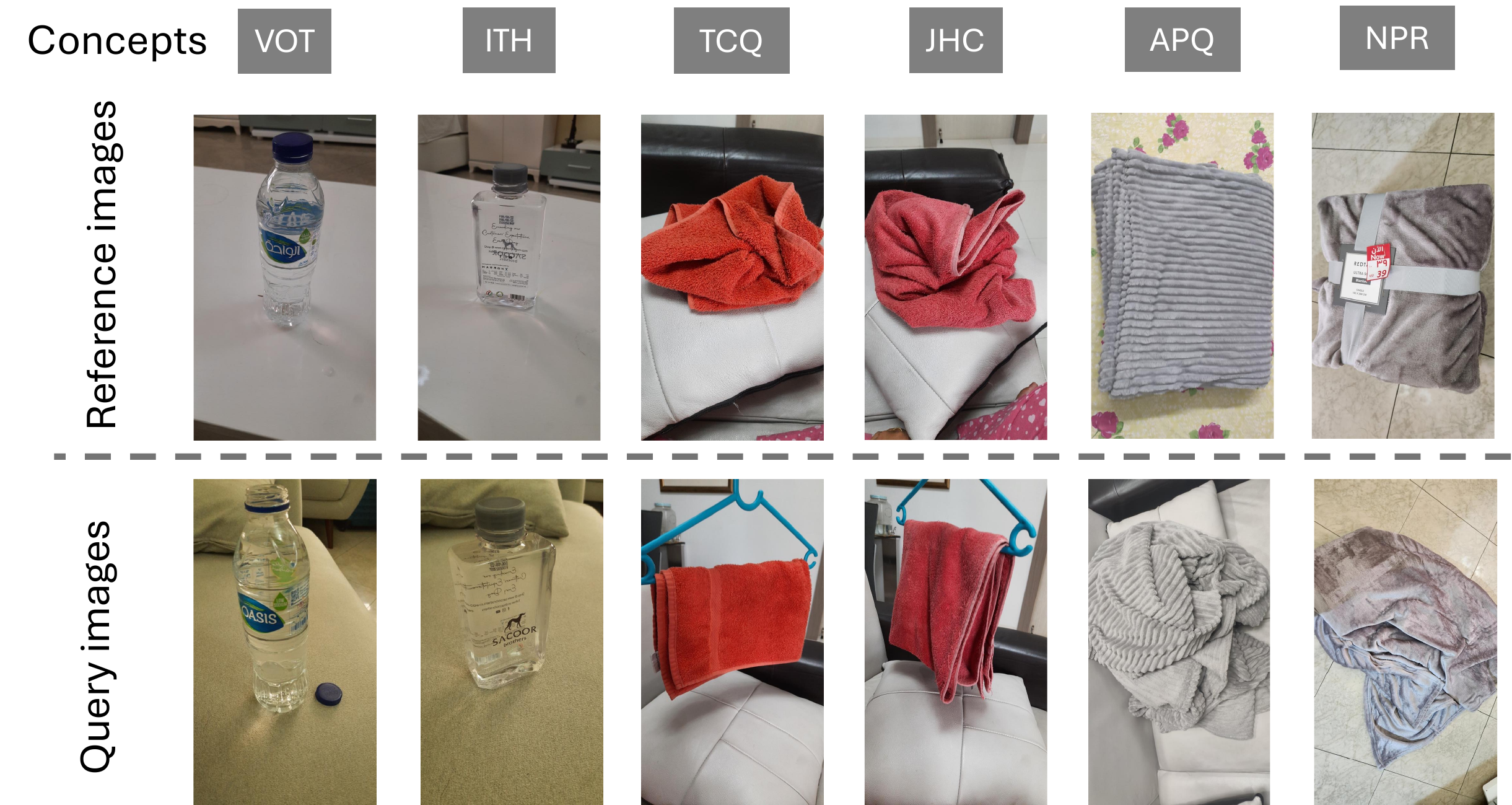}
    \caption{Qualitative visualization of concepts for the proposed \textit{\datasetshort{}} dataset. In order, samples from bottles, towels and clothes. \textbf{Top:} reference images for personalization with their concept indicated above. \textbf{Bottom:} query images at inference time.}
    \label{fig:dataset-viz}
\end{figure}

\section{\datasetshort{} dataset} 
\label{dataset}
Current personalization datasets~\citep{alaluf2024myvlm, nguyen2024yollava} focus on a limited number of concepts ($\leq$ 50) from a small set of categories. Often, the personal concepts feature objects with highly distinctive traits, \eg a pair of sporty shoes in pink and orange color. The images containing the personal concepts are also well posed in images with goof lighting. As such they exhibit limited recognition challenge as there is little visual ambiguity among personal concepts. Other datasets focus on synthetically generated personal concepts~\citep{hao2025remember}, which may not accurately reflect the challenges encountered in real-world scenarios.

We thus introduce a more challenging benchmark for VLM personalization, \textit{\dataset{} (\datasetshort{})}, comprising many everyday object categories and each category having multiple instances with similar visual features. 
We construct \datasetshort{} by repurposing a publicly available dataset, originally proposed in~\citep{sarkar2024learning} to study robust object recognition and retrieval. This dataset contains images featuring everyday objects belonging to the same category, which are observed in different poses, viewpoints and possibly different states, \eg, t-shirt being folded or hang.
Specifically, we restructure the existing dataset by creating new splits. Ideally, the query images at inference time should be different from the reference images used for model personalization.
To this end,
we consider the images associated to a concept $\conceptname_i$, and compute the average image embedding representation using a pre-trained image encoder~\citep{radford2021learning}. 
This average image embedding serves as an anchor embedding to determine which samples can be used as reference images for model personalization, and which are reserved for inference.
Specifically, as the reference image, we select the image whose visual embedding is closest to the average embedding corresponding to a concept. as query images, we select images whose embeddings are the furthest from the average one. 

With such splitting strategy, we create a challenging evaluation dataset for personal concept recognition. At inference time, specific concepts may appear in different poses, lighting conditions, and contexts. 
In total, \datasetshort{} contains 329 personal concepts, spanning 21 categories of everyday entities such as clothes, bottles, trolleys and umbrellas. Among existing datasets \citep{nguyen2024yollava, alaluf2024myvlm}, \datasetshort{} presents a larger number of concepts per category ranging from 2 (headphones) to 69 (retail) concepts and stress-tests personalization models in well-known challenging settings, namely deformation for non-rigid objects, different illumination conditions and different object states, \eg, folded clothes. We provide additional details on the \datasetshort{} dataset in the supplementary.
Fig.~\ref{fig:dataset-viz} showcases reference images for personalization and query images. While this work focuses on a single reference image for personalization, our dataset construction strategy can be extended to multiple reference images and other datasets. 

\section{Experiments}
\label{sec:experiments}
We evaluate \methodshort{} on established personalization benchmarks, \ie MyVLM~\citep{alaluf2024myvlm} and \yollava~\citep{nguyen2024yollava}, as well as our newly introduced \datasetshort{} dataset. 
We first outline our evaluation protocol and the implementation details, then discuss our results comparing \methodshort{} with state-of-the-art personalization methods.
Finally, we present extensive ablation studies that highlights the importance of the design choices of our method.

\subsection{Experimental setup}
\noindent\textbf{Datasets.}
In addition to \datasetshort{}, we also conduct evaluation on existing personalization benchmarks, \ie MyVLM~\citep{alaluf2024myvlm} and \yollava. MyVLM~\citep{alaluf2024myvlm} considers 29 personal concepts, while Yo'LLaVa~\citep{nguyen2024yollava} includes images of 40 concepts, including objects, people, and buildings. Unlike the original evaluation protocol~\citep{alaluf2024myvlm, nguyen2024yollava}, which uses multiple positive and negative samples for training, our approach relies on a single reference image during personalization and does not require any negative samples. This setting, while being more challenging, better aligns with real-world use cases, as user input for personalization is minimized. 

\noindent\textbf{Downstream tasks and Metrics.} Following prior works~\citep{nguyen2024yollava, hao2025remember}, we evaluate \methodshort{} on three tasks: object recognition, captioning and personalized VQA. In the recognition task, the model determines whether a specific concept of interest is present in a query image. Positive samples correspond to images containing the personal concept of interest, while negative samples are images featuring other concepts. We frame object recognition as a binary classification problem and report recall (\texttt{Pos.~Acc.}), specificity (\texttt{Neg.~Acc.}), and their uniformly weighted average (\texttt{Wtd}). 
On the captioning task, instead, we assess Hard Recall (\texttt{Recall}), where the model should detect any personal concept of interest without prior knowledge of the specific concept. This is measured by the fraction of times the concept name appears in the model’s generated captions for its test images.
Finally, on personalized VQA, we evaluate the model’s ability in answering closed-set questions about the personalized concepts, measuring the overall answering accuracy. Results are reported as the average over three different seeds.

\noindent\textbf{Implementation Details.} We use Mini-CPM-o-2.6 \cite{minicpm} as the underlying \vlm for all experiments, chosen for its lightweight design and strong comparitive image understanding capabilities. Built on LLaVA-UHD, it employs SigLIP \cite{siglip} as the vision encoder, LLaMA 3.1 as the text decoder, and a multimodal projection layer for alignment.

To extract fingerprint attributes and descriptions, we prompt Mini-CPM-o-2.6 with task-specific templates (see supplementary for full details). For retrieval, we use FAISS~\cite{faiss} following~\cite{hao2025remember}, and construct the personal database using CLIP ViT-L/14-336 to encode both image and caption embeddings. We set $K{=}3$ in all experiments to balance accuracy and efficiency.

On MyVLM and Yo'LLaVA datasets, we crop images before encoding to isolate individual concepts, given the presence of multiple objects per image.
\subsection{Comparison with the state of the art}  
\paragraph{Baselines.} We compare \methodshort{} against state-of-the-art methods for \vlm personalization, including MyVLM~\citep{alaluf2024myvlm} and \yollava{}~\citep{nguyen2024yollava}, which require training to adapt to new personalized concepts, as well as RAP~\citep{hao2025remember}, a retrieval-based approach with a \vlm instruction-tuned for personalized responses. In addition, we consider prompting-based strategies evaluated in~\cite{nguyen2024yollava}: \texttt{GPT-4V~+~Vprompt}, which is provided with both query and reference images for direct comparison; LLaVA~\citep{liu2023visual}, which relies solely on the pre-trained model with no personalized information; and \texttt{LLaVA~+~prompt}, where the model is prompted with a general description of the personal concept. We further evaluate \texttt{MiniCPM-o~+~prompt}~\citep{yu2024rlaif}, a retrieval-based reasoning approach where the \vlm infers the presence of a concept based on the query image and descriptions of the closest retrieved elements. 
To ensure fair comparison, we also include \texttt{Yo'LLaVA w/ MiniCPM-o 2.6} as a baseline where we train the \yollava{} algorithm replacing its VLM to MiniCPM-o 2.6.

\noindent\textbf{Results on Recognition and Captioning.} Table~\ref{tab:recognition_accuracy_datasets} reports recognition and captioning results on standard personalization benchmarks.
On MyVLM, \methodshort{} achieves the best weighted accuracy (97.4\%) and competitive negative accuracy, on par with RAP. Interestingly, \texttt{MiniCPM-o + prompt} attains the highest positive accuracy but performs poorly on negative accuracy, suggesting overconfidence in answering “yes” regardless of query context.

On Yo'LLaVA, \methodshort{} achieves the highest captioning recall (87.1\%), outperforming the next best by 5.5\%. It also strikes a stronger balance in recognition, improving negative accuracy over the naive prompting baseline by +4.8\%. Compared to RAP, \methodshort{} shows better generalization (+2.2\% Wtd, +5.1\% Pos.~Acc.), despite being training-free.

Furthermore, \methodshort{} surpasses \texttt{Yo'LLaVA w/ MiniCPM-o 2.6}, confirming that the gains stem from our reasoning-based approach rather than backbone improvements. Overall, \methodshort{} offers a favorable trade-off between positive and negative accuracy, demonstrating the strength of our attribute-based verification. Captioning performance of MyVLM and \yollava{} is notably lower, as both models rely on the concept name being explicitly provided in the prompt. Without access to this information, they often fail to generate captions that correctly reference or ground the intended concept.

On our challenging \datasetshort{} dataset, \methodshort{} continues to show strong recognition and captioning performance, driven by its use of fingerprint attributes and multimodal reasoning. It outperforms all baselines on three of four metrics, clearly outperforming the unimodal \texttt{MiniCPM-o + prompt} baseline.

On recognition, \methodshort{} achieves 91.8\% weighted accuracy, surpassing RAP by +2.8\%. Compared to training-based models like MyVLM (+29.6\%) and \yollava{} (+19.8\%), the gap widens further, highlighting the difficulty of learning robust concepts under limited reference images.

Similar trends are seen in captioning: \methodshort{} achieves 72.5\% recall, outperforming RAP (64.1\%) and \texttt{MiniCPM-o + prompt} (65.7\%), where training-based models struggle with visually similar objects. These results reinforce the effectiveness of our training-free, attribute-guided strategy.

\begin{figure}[t!]
    \centering
    \includegraphics[width=\linewidth]{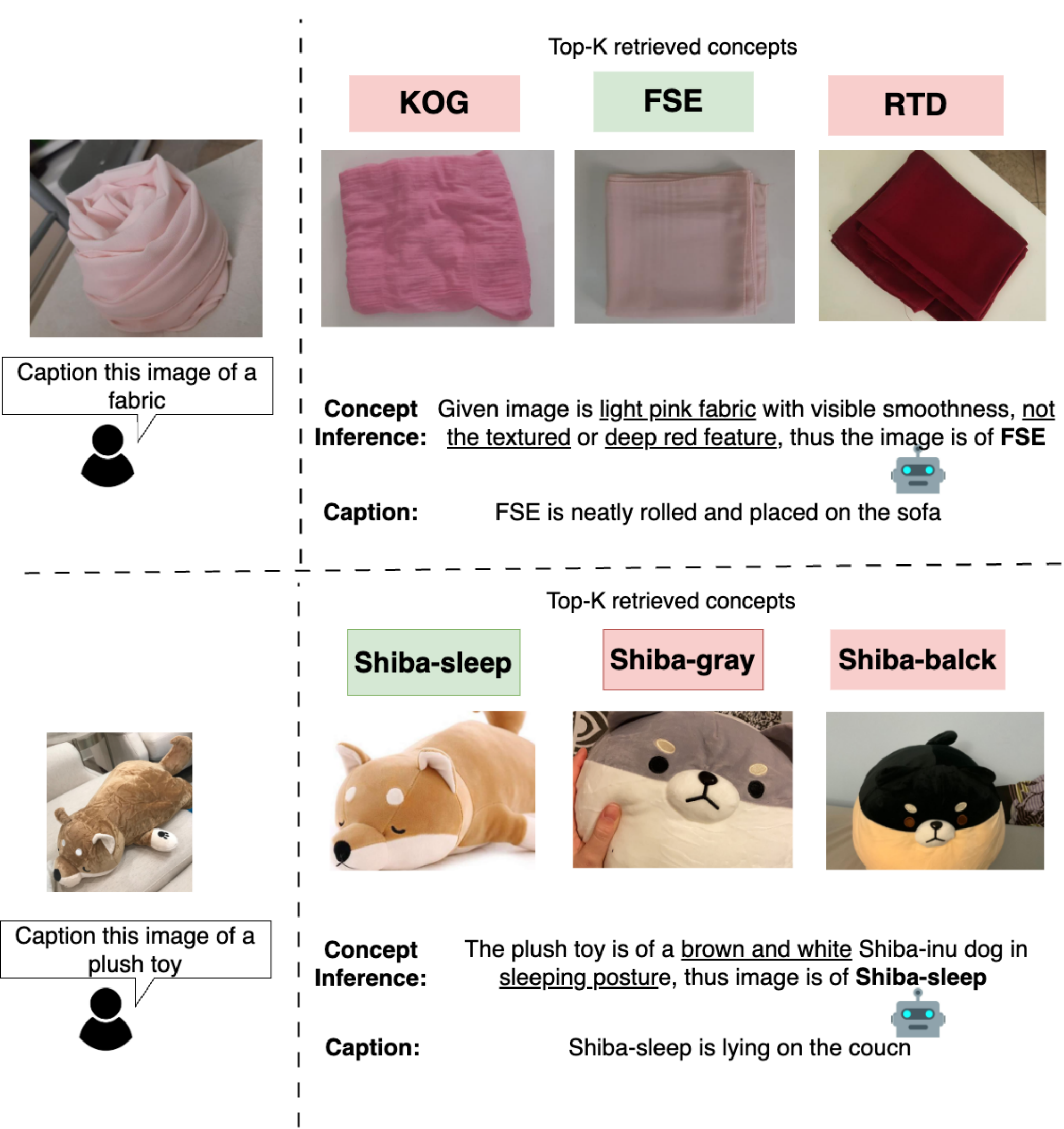}
    \caption{Qualitative examples. Given a query image and a user prompt (left), \methodshort{} retrieves the most similar Top-K concepts, analyzes a set of fingerprint attributes, 
    and generates a precise, personalized caption. The key attributes enabling the model to recognize the correct concept name (bold) are underlined for clarity.}
    \label{fig:qual_result}
\end{figure}

In Figure \ref{fig:qual_result}, we showcase examples of personalization. The top row displays an image of fabric from our \datasetshort{} dataset, while the bottom row features an image from the \yollava dataset. Both examples illustrate how \methodshort{} accurately identifies the correct concept name for each object by reasoning over its fingerprint attributes, despite the presence of very similar candidate alternatives.
\begin{table}[t]
\centering
\resizebox{\linewidth}{!}{
\begin{tabular}{lccccc}
\toprule
\multirow{3}{*}{\textbf{Method}} 
& \multicolumn{3}{c}{\textbf{Recognition}} & \textbf{Captioning}\\
\cmidrule(lr){2-4}
\cmidrule(lr){5-6}
 & Pos. Acc. & Neg. Acc. & Wtd & Recall \\ 
\midrule
\multicolumn{6}{c}{\cellcolor{Gray!16}\textbf{MyVLM Dataset~\citep{alaluf2024myvlm}}}\\
MyVLM~\citep{alaluf2024myvlm} & 96.6 & 90.9 & 93.8  & 0.02\\
\yollava~\citep{nguyen2024yollava} & \underline{97.0} & 95.7 & 96.4  & 0.1 \\
RAP~\citep{hao2025remember}  & 94.4 & \textbf{98.8} & \underline{96.6}  & \underline{88.0}\\
MiniCPM-o + prompt~\citep{yu2024rlaif} & \textbf{ 98.5} & 93.7  & 96.1  & 87.4 \\
\methodshort{} \textbf{(Ours)} & 96.3 & \underline{98.47} & \textbf{97.4}  & \textbf{91.4}  \\
\midrule
\multicolumn{6}{c}{\cellcolor{Gray!16}\textbf{\yollava Dataset~\citep{nguyen2024yollava}}}\\
\yollava~\citep{nguyen2024yollava} & \textbf{94.9} & 89.8 & \underline{92.4}  & 0.2\\
\yollava w/ MiniCPM-o 2.6 & 62.8 & 62.2 & 62.5 &- \\
GPT-4V+ Vprompt~\citep{nguyen2024yollava} & 80.9 & \textbf{99.2} & 90.1  & -\\
RAP~\citep{hao2025remember}  & 86.0 & \textbf{99.2} & 92.2  & \underline{79.7}\\
MiniCPM-o + prompt~\citep{yu2024rlaif} & \underline{91.2} & 93.5 & \underline{92.4}  & 73.9\\
\methodshort{} \textbf{(Ours)} & 91.1 & 97.7 & \textbf{94.4}  & \textbf{87.1} \\
\midrule
\multicolumn{6}{c}{\cellcolor{Gray!16}\textbf{\datasetshort{} Dataset}}\\
MyVLM~\citep{alaluf2024myvlm} & 66.0 & 58.5 & 62.2 & 0.3\\
\yollava~\citep{nguyen2024yollava} & 75.1 & 69.0 & 72.0 & 6.6 \\
RAP~\citep{hao2025remember} & \textbf{92.9} & 85.2 & \underline{89.0} & 64.1 \\
MiniCPM-o + prompt~\citep{yu2024rlaif} & 73.0 & \underline{92.5} & 82.7 & \underline{65.7} \\
\methodshort{} \textbf{(Ours)} & \underline{90.2} & \textbf{92.8} & \textbf{91.8} & \textbf{72.5}  \\
\bottomrule
\end{tabular}
}
\caption{Recognition and captioning performance ($\uparrow$) on established personalization benchmarks and the newly proposed \datasetshort{}. 
}
\label{tab:recognition_accuracy_datasets}
\end{table}



\noindent\textbf{Results on personalized VQA.} Table~\ref{tab:yollava-vqa} reports the performance on personalized VQA on \yollava{} dataset. We follow a similar evaluation protocol as in \citep{nguyen2024yollava} and compare against baselines and competitors (results reported in the original paper). While \texttt{LLaVA + prompt}, \yollava{} and RAP achieves similar performance, \methodshort{} scores the highest accuracy with a large improvement with respect to current state-of-the-art (+3.3\% compared to RAP). \yollava and RAP training strategies rely on assumptions about the possible questions to be asked. Instead, \methodshort{} imposes no priors, allowing any questions that might arise during inference time. Compared to other training-free baselines based on LLaVA and GPT-4V, \methodshort{} more accurately identifies. 

\begin{table}[t]
\centering
\resizebox{0.6\linewidth}{!}{
\begin{tabular}{lc}
\toprule
\textbf{Method} & \textbf{Accuracy} \\
\midrule
Yo'LLaVA~\citep{nguyen2024yollava} & 92.9 \\
RAP~\citep{hao2025remember}& \underline{93.2} \\
GPT-4V + Vprompt~\citep{nguyen2024yollava} & 86.6\\
LLaVA~\citep{nguyen2024yollava}  & 89.9 \\
LLaVA + prompt~\citep{nguyen2024yollava} & 92.5 \\
\methodshort{} \textbf{(Ours)}  & \textbf{96.5} \\
\bottomrule
\end{tabular}
}
\caption{VQA performance in terms of answering accuracy ($\uparrow$) in \yollava~\citep{nguyen2024yollava} Dataset.}
\label{tab:yollava-vqa}
\end{table}

\subsection{Ablation Study}
In this section, we analyze the impact of key components of our method, including the role of pairwise reasoning, the introduction of fingerprint attributes, and 
\vlm's CoT reasoning (\cref{tab:ablation_inputs}). Then, we ablate the cross-modal attribute verification strategy in comparison with recent alternatives (\cref{tab:verification}). Finally, we evaluate the role of different embeddings for the concept retrieval (\cref{tab:retrieval_embeddings}). All experiments are conducted on the most challenging dataset, \datasetshort{}. 

\noindent\textbf{Impacts of main components.} Tab.~\ref{tab:ablation_inputs} presents the ablation on the main components of our proposed method on the recognition and captioning task. The first row reports results when we naively prompt the \vlm to output the best-matched concept, given only the query image and the textual information of all candidate concepts, without CoT reasoning, without the use of fingerprint attributes, and without pairwise reasoning. Its performance on the recognition and captioning tasks are greatly limited, when compared to the sixth row (our full method~\methodshort{}).
Interestingly, when only activating CoT reasoning without fingerprint attributes (second row), the performance in recognition worsens compared to that of the first row. Yet, CoT reasoning tied with fingerprint attributes (third row), greatly improves the performance. 

The fourth row reports results when only the computationally expensive pairwise reasoning is activated. Even without the use of fingerprint attributes, the recognition performance is already better than our method~\methodshort{} at the sixth row, while its captioning performance remains limited. When CoT reasoning is leveraged together with pairwise reasoning without considering fingerprint attributes (the fifth row), we observe a slight performance gain in captioning, despite a minor drop in recognition.
Eventually, the sixth row reports the final performance of \methodshort{}, obtaining the best performance in both recognition and captioning task. 
To investigate VLM-generated fingerprint attributes, we also report results in the last row when a human-specified pre-defined set (\eg, color, shape, pattern, printed text) is provided during the database creation. Such attributes leverage privileged human knowledge to improve fine-grained recognition. Interestingly, \methodshort{} is almost on par with it, confirming that the VLM-generated fingerprint attributes are indeed informative and distinctive.

\begin{table}
\centering
\resizebox{0.95\linewidth}{!}{  
\begin{tabular}{ccc ccc}
\toprule  
\makecell{\textbf{Pairwise} \\\textbf{Reasoning}} & \makecell{\textbf{Fingerprint}\\\textbf{Attributes}} & \makecell{\textbf{Reasoning}\\\textbf{CoT}} &  \makecell{\textbf{Recognition}\\\textbf{Wtd}} & \makecell{\textbf{Captioning} \\ \textbf{Recall}} \\  
\midrule  
\textcolor{red}{\normalsize \ding{55}} & \textcolor{red}{\normalsize \ding{55}} & \textcolor{red}{\normalsize \ding{55}} & 86.5 & 62.2\\
\textcolor{red}{\normalsize \ding{55}} & \textcolor{red}{\normalsize \ding{55}} & \textcolor{Green}{\normalsize \ding{51}} & 84.7 & 62.8\\
\textcolor{red}{\normalsize \ding{55}} & \textcolor{Green}{\normalsize \ding{51}} & \textcolor{Green}{\normalsize \ding{51}} & 91.8 & \underline{71.2}\\
\textcolor{Green}{\normalsize \ding{51}} & \textcolor{red}{\normalsize \ding{55}} & \textcolor{red}{\normalsize \ding{55}} & 92.3  & 65.9\\
\textcolor{Green}{\normalsize \ding{51}} & \textcolor{red}{\normalsize \ding{55}}  &  \textcolor{Green}{\normalsize \ding{51}} & \underline{91.6} & 67.3\\
\cellcolor{method-green}\textcolor{Green}{\normalsize \ding{51}} & \cellcolor{method-green} \textcolor{Green}{\normalsize \ding{51}}  & \cellcolor{method-green}\textcolor{Green}{\normalsize \ding{51}} & \cellcolor{method-green}\textbf{91.8} & \cellcolor{method-green}\textbf{72.5} \\

\cellcolor{tab-red}\textcolor{Green}{\normalsize \ding{51}} & \cellcolor{tab-red}privileged  & \cellcolor{tab-red}\textcolor{Green}{\normalsize \ding{51}} & \cellcolor{tab-red}92.5 & \cellcolor{tab-red}72.8 \\  
\bottomrule  
\end{tabular}
}
\caption{Ablation on pairwise reasoning, fingerprint attributes, and the use of CoT reasoning for \inlineColorbox{method-green}{\methodshort{}} in terms of weighted recognition metrics ($\uparrow$) and captioning recall ($\uparrow$) on \datasetshort{}. We include a \inlineColorbox{tab-red}{privileged} version of our approach with human pre-defined fingerprint attributes for concepts.}
 \label{tab:ablation_inputs}
\end{table}

\noindent\textbf{Verification strategies.}  
We compare the proposed cross-modal attribute verification strategy with other approaches accounting for VLM uncertainty. In particular, we consider \texttt{abstention} where the model is prompted with the instruction to output ``I am not sure'' for uncertain cases, and \texttt{logits}-based evaluation considering the logits associated to the ``I don't know'' option from the set of possible answers. For completeness, we show results when retrieved elements are entirely processed via the more demanding pairwise reasoning (\texttt{Pairwise-reasoning}) and when no verification step is performed, \ie, never triggering the pairwise reasoning, (\textit{No estimation}). Table~\ref{tab:verification} shows that our multimodal verification strategy achieves a recall of 72.5\%, outperforming abstention (70.7\%) and no estimation (71.2\%) and providing additional gain with respect to the more demanding pairwise reasoning only. Our verification mechanism balances accuracy and efficiency, effectively reducing hallucinations while maintaining competitive performance.  
\begin{table}
\centering
 \resizebox{0.62\linewidth}{!}{  
\begin{tabular}{lc}
\toprule  
\textbf{Method} & \textbf{Captioning Recall} \\
\midrule
Pairwise-reasoning & \underline{72.3} \\
\hdashline
No estimation & 71.2 \\
Abstention & 70.7 \\
Logits-based & 70.9 \\
\cellcolor{method-green}Attr. Verification \textbf{(Ours)} & \cellcolor{method-green} \textbf{72.5}\\
\bottomrule  
\end{tabular}
}
\caption{Ablation on verification strategies for \inlineColorbox{method-green}{\methodshort{}} on \datasetshort{} dataset. Performance is evaluated based on captioning recall ($\uparrow$).} 
\label{tab:verification}
\end{table}

\noindent\textbf{Retrieval.}  
We asses the impact of different retrieval strategies for concept retrieval in \methodshort{}, as evaluated in terms of HIT@K metric measuring how often the personalized concept of interest is correctly retrieved in the top-K elements. We ablate different options of $K={1, 3, 5, 10}$. In particular, we compare representations from different pre-trained models such as DINOv2~\citep{oquab2023dinov2} and CLIP~\citep{radford2021learning}. For CLIP we consider both text and visual embeddings, denoted as CLIP Image and CLIP Text, respectively. For multimodal retrieval, we also ablate a two-step approach where concepts are first retrieved based on visual embeddings, and then re-ranked by relying on textual similarity. 
As shown in Table~\ref{tab:retrieval_embeddings}, DINOv2 achieves the highest H@1 performance (68.5\%), while the proposed strategy in \methodshort{} fusion performs best when considering the Top-3 and Top-5 retrieved items with 87\% and 93\%, respectively. These results highlight the benefits of combining image and text embeddings for robust retrieval. 

\begin{table}[t]
\centering

\resizebox{0.8\columnwidth}{!}{%
\begin{tabular}{lcccc}
\toprule
\textbf{Embedding} & \textbf{H@1} & \textbf{H@3} & \textbf{H@5} & \textbf{H@10} \\
\midrule
DINOv2 & \textbf{68.5} & 83.9 & 90.2 & \textbf{96.4} \\
CLIP-Image & 54.0 & 76.9 & 86.0 & 92.9 \\
CLIP-Text & 60.6 & 80.7 & 89.0 & 93.6 \\
Multimodal (2-step) & 65.0 & \underline{84.7} & \underline{90.6} & 93.6 \\
\cellcolor{method-green}\methodshort{} \textbf{(Ours)} & \underline{67.5} & \textbf{87.0} & \textbf{93.0} & \underline{96.3} \\
\bottomrule
\end{tabular}%
}
\caption{Performance with different retrieval strategies evaluated in terms of HIT@K ($\uparrow$) on \datasetshort{} dataset.}
\label{tab:retrieval_embeddings}
\end{table}

\section{Conclusions}
\label{sec:conclusion}
We explored training-free personalization of \vlms and proposed \methodshort{}, a novel approach that uses pre-trained VLMs to retrieve and reason over textual fingerprint attributes. \methodshort{} uniquely identifies user-specific concepts even in visually ambiguous scenarios. We demonstrated its effectiveness through extensive experiments on standard benchmarks and our new \datasetshort{} dataset, achieving state-of-the-art performance in personalized concept recognition and captioning. Future work will aim to reduce computational overhead and improve inference in cluttered scenes with similar-looking concepts. 
\section{Acknowledgments}
This work was supported by the PNRR ICSC National Research Centre for HPC, Big Data and Quantum Computing (CN00000013), IPCEI Cloud (DM 27 June 2022 – IPCEI-CL-0000007) from the Italian Ministry of Enterprises and Made in Italy, and FAIR – Future AI Research (PE00000013), funded by the EU’s NextGeneration initiative. Additional support was provided by the EU projects ELIAS (No. 101120237) and ELLIOT (No. 101214398). We acknowledge ISCRA for awarding this project access to the LEONARDO supercomputer, owned by the EuroHPC Joint Undertaking, hosted by CINECA (Italy). We acknowledge EuroHPC Joint Undertaking for awarding us access to MareNostrum5 as BSC, Spain.

{
    \small
    \bibliographystyle{ieeenat_fullname}
    \bibliography{main}
}

\clearpage
\setcounter{page}{1}
\maketitlesupplementary
\section{Supplementary Overview}
In this supplementary, we provide a detailed statistics of the PerVA dataset in Section \ref{supp:perva data stats}, qualitative example \methodshort{} demonstrating our proposed Attribute Focused CoT and Pairwise Image Matching in Section ~\ref{supp:additional-qual}, details on prompting in Section \ref{supp:prompts} and analysis of the proposed \methodshort{} on MyVLM and \yollava datasets in Section Sec.~\ref{supp:additional-results}.

\subsection{PerVA Dataset Overview and distribution}
\label{supp:perva data stats}

The PerVA dataset comprises 21 object categories with a total of 67,482 images distributed across 329 unique concepts. The dataset is structured to support training-free retrieval-augmented generation (RAG) approaches, with each concept represented by multiple training images and a smaller set of test images.

The dataset follows a balanced concept distribution across train and test splits, with identical concept coverage (329 concepts each) but different image quantities per concept. The training split contains 59,392 images with an average of 180.5 images per concept, while the test split contains 8,090 images with an average of 24.6 images per concept.

For our RAG-based approach, we constructed a reference database containing exactly one representative image per concept (329 images total), ensuring complete concept coverage across all categories.

\begin{table}[h!]
\centering
\label{tab:perva_stats}
\resizebox{0.95\linewidth}{!}{
\begin{tabular}{lcccc}
\toprule

\textbf{Category} & \textbf{Concepts} & \textbf{Train Images} & \textbf{Test Images} & \textbf{DB Images} \\
\midrule
decoration & 62 & 12,935 & 2,011 & 62 \\
retail & 69 & 12,678 & 1,750 & 69 \\
clothe & 30 & 5,437 & 873 & 30 \\
bag & 25 & 4,534 & 426 & 25 \\
veg & 20 & 3,419 & 604 & 20 \\
plant & 17 & 2,971 & 485 & 17 \\
toy & 13 & 2,440 & 136 & 13 \\
book & 12 & 2,018 & 140 & 12 \\
cup & 11 & 1,691 & 396 & 11 \\
pillow & 9 & 1,584 & 174 & 9 \\
tumbler & 8 & 1,496 & 296 & 8 \\
bowl & 7 & 1,186 & 83 & 7 \\
towel & 7 & 1,138 & 83 & 7 \\
remote & 7 & 1,047 & 79 & 7 \\
plate & 7 & 1,013 & 127 & 7 \\
bottle & 7 & 978 & 102 & 7 \\
tie & 6 & 885 & 53 & 6 \\
tro\_bag & 5 & 912 & 169 & 5 \\
umbrella & 3 & 536 & 65 & 3 \\
headphone & 2 & 259 & 22 & 2 \\
telephone & 2 & 235 & 16 & 2 \\
\midrule
\textbf{Total} & \textbf{329} & \textbf{59,392} & \textbf{8,090} & \textbf{329} \\
\bottomrule
\end{tabular}
}
\caption{Category-wise distribution of the PerVA dataset}
\end{table}

The dataset exhibits natural category imbalance reflecting real-world object distributions, with decoration and retail categories containing the most concepts (62 and 69, respectively), while telephone and headphone categories contain the fewest (2 each). This distribution provides a realistic testbed for evaluating retrieval-based recognition systems across diverse object categories with varying levels of intra-category diversity.
\subsection{Additional qualitative example}
\label{supp:additional-qual}
\begin{figure*}[t]
    \centering
    \includegraphics[width=\textwidth]{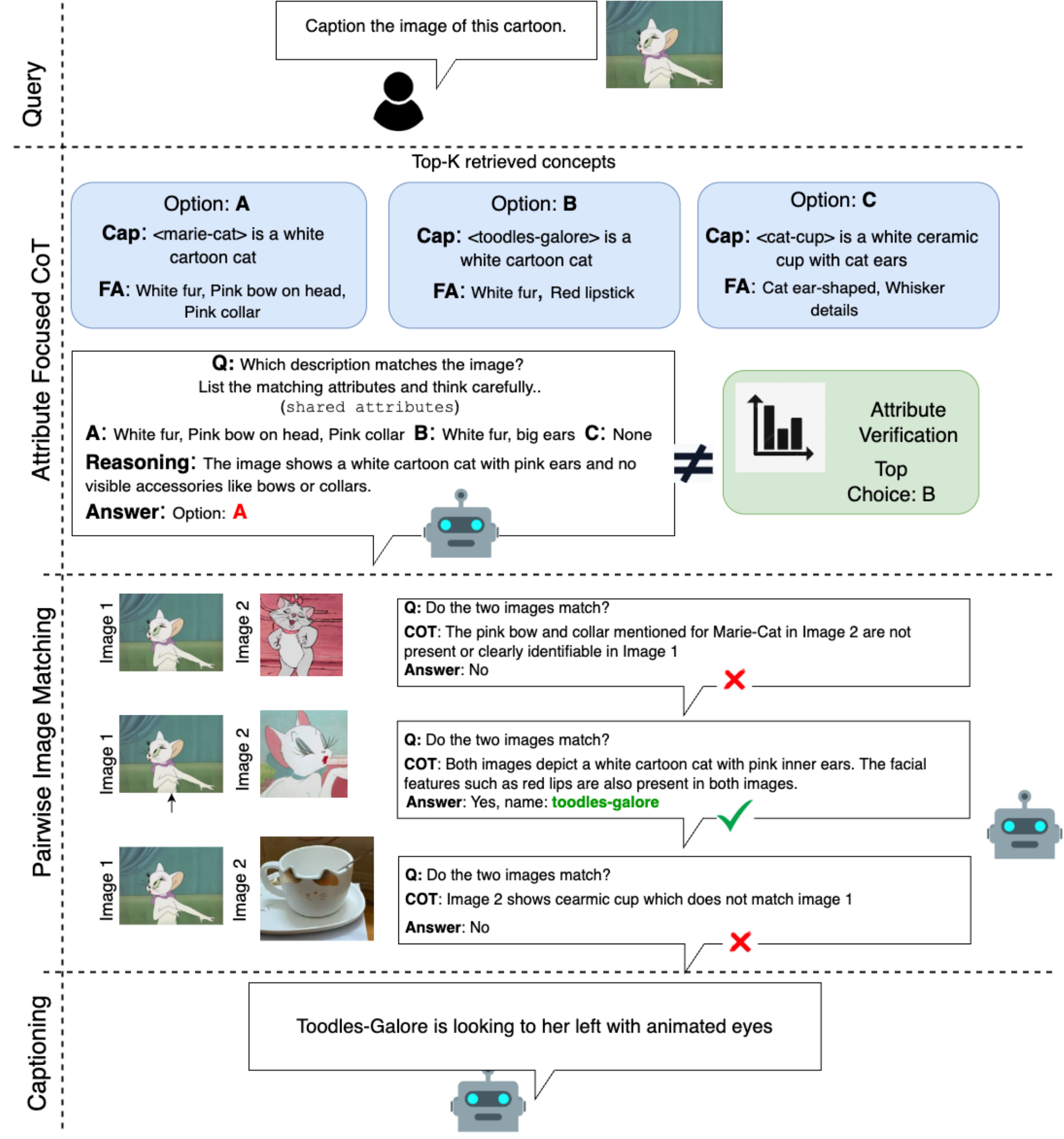} 
    \caption{Qualitative example of the Concept Inference with Retrieval-Reasoning of \methodshort{}}
    \label{fig:supp_qual}
\end{figure*}

In Figure \ref{fig:supp_qual}, we qualitatively demonstrate the Concept Inference with Retrieval-Reasoning of \methodshort{}. Given a query about a personalized object (in this case, a cartoon cat) and the Top-K retrieved concepts along with their descriptions, \methodshort{} first performs attribute-focused Chain-of-Thought (CoT) reasoning on the fingerprint attributes of each retrieved concept. The model is instructed to compare the unique attributes between the query image and each description, then reason over them to identify the correct concept name. However, in this case, the model hallucinated the attribute 'pink bow on head' for the query, which is a fingerprint attribute for option A (i.e., marie-cat), leading to misclassification. Since attribute verification failed here, \methodshort{} proceeds to perform more computationally expensive pairwise reasoning, correctly identifying the concept name, which it then uses to generate a personalized caption.\\

\subsection{Prompting}
\label{supp:prompts}
In this section, we define the prompts used for our personalization task. Figure~\ref{fig:prompt cache generation} shows the prompt template for creating the personal database. Here, we provide the model with the image, object category, and concept name, prompting it to identify the distinct attributes that make the object unique, compared to similar objects in the same category.

Figure~\ref{fig:prompt cot reasoning} illustrates the prompt template for attribute-focused CoT reasoning. The model is shown the query image along with the retrieved concepts from the database. For each retrieved concept, the \vlm is asked to compare the query image with its description and list the attributes that are common between the two. If no shared attributes are found, the model may omit mentioning any. Based on the matching attributes, the model carefully selects the option that best describes the query image.

Finally, Figure~\ref{fig:prompt pairwise comparison} shows the prompt for multimodal pairwise reasoning. The model is provided with the query image, the concept reference image, and the associated description. The model is prompted to compare the two images and determine whether they match. The description of the reference image is included to help the model identify relevant attributes in the query image, enabling a more informed decision. Please note that the prompts shown are for demonstration purposes only. In practice, we populate the descriptions of retrieved objects directly from the database.

\begin{figure}[t]
    \centering
\small 
\begin{myquote}
Describe the $<g_i>$ in the image identified by the concept-identifier
$<c_i>$ and highlight what makes it unique.\\
Your response MUST be in valid JSON format and must follow EXACTLY the format below:\\
\{\\
    general: a brief description of the object in one sentence.,\\
    category: category of the object,\\
    distinct features: [List of distinct features that makes the object unique],\\
\}\\
IMPORTANT:\\
- List only the most distinguishing features that set this object apart.\\
- Avoid generic descriptions that apply to every object in this category.\\
- Do not include any extra text or commentary.
Any deviation from this format will be considered incorrect.
\end{myquote}
    \caption{Prompt Template for Personal Database Creation}
    \label{fig:prompt cache generation}
\end{figure}

\begin{figure}[t]
\centering
\small 
\begin{myquote}
You are a helpful AI agent specializing in image analysis and object recognition.\\
Your task is to analyze and compare a query image with three provided descriptions.\\
Below are the description(s)\\

A. Name: $<{c_{i_1}}>$, \\
Info: \{general: A generic description about $<c_{i_1}>$,\\ 
category: category of $<c_{i_1}>$,\\
distinct features: [distinct feature 1, distinct feature 2, ...]\}\\

B. Name: $<c_{i_2}>$,\\ 
Info: \{general: A generic description about $<c_{i_2}>$,\\ 
category: category of $<c_{i_2}>$,\\
distinct features: [distinct feature 1, distinct feature 2, ...]\}\\

C. Name: $<c_{i_3}>$,\\ 
Info: \{general: A generic description about $<c_{i_3}>$,\\ 
category: category of $<c_{i_3}>$,\\
distinct features: [distinct feature 1, distinct feature 2, ...]\}\\

Your Task:\\
- Compare the query image with each description and answer the following question:\\      
Which description matches the subject in the image?\\ Answer in A, B, C.\\            
- List shared attributes between the image and each description very concisely\\          
- If no attributes match for a certain, generate None\\        
- Provide a brief reasoning for your final answer.\\       
- Respond strictly in the following JSON format:\\
\{\\
"A": "[Matching attributes for option A]",\\ 
"B": "[Matching attributes for option B]",\\
"C": "[Matching attributes for option C]",\\ 
"Reasoning": "$<$Brief justification$>$",\\
"Answer": "$<$one of A, B, C$>$"\\
\}\\

Any deviation from this format will be considered incorrect. \\
Do not output any additional text.\\
\end{myquote}

\caption{Prompt template for Attribute-focused CoT reasoning}
\label{fig:prompt cot reasoning}
\end{figure}

\begin{figure}[t]
    \centering
    \small 
    \begin{myquote}

    You are a helpful AI agent specializing in image analysis and object recognition.\\
    You are given two images: **Image 1** and **Image 2**.\\ 
    Additionally, the name and a textual description of the object in **Image 2** 
    is also provided below:\\
    
    1. Name: $<c_i>$,\\ 
    Info: \\
    \{general: A generic description about $<c_i>$,\\ 
    category: category of $<c_i>$,\\ 
    distinct features: [distinct feature 1, distinct feature 2, ...]\}\\
    Task:\\
    - Compare the two images and answer the following question.\\
    Can you see $<c_i>$ in this Image 1?\\ 
    Answer with a single word, either yes or no.\\
    - Provide your reasoning based on the two images and the given description.\\
    - Generate your response with JSON format:\\
    \{\\
      "Reasoning": "$<$Your reasoning in 2-3 sentences.$>$",\\
      "Answer": "$<$yes or no$>$"\\
    \}\\
    Output only the JSON response. DO NOT output any additional text.\\
    \end{myquote}
    \caption{Prompt Template for Image based pairwise comparison}
    \label{fig:prompt pairwise comparison}
\end{figure}

\subsection{Additional Results}
\label{supp:additional-results}
In this section we report additional ablation analysis on existing personalization datasets in the literature, namely MyVLM~\citep{alaluf2024myvlm} and \yollava~\citep{nguyen2024yollava}. 

Tab.~\ref{tab:ablation_myvlm} and Tab.~\ref{tab:ablation-yollava} reports the ablation analysis of the proposed approach on the use of pairwise-reasoning, fingerprint attributes and CoT reasoning of the underlying \vlm. Results are mostly consistent with the main paper results on \datasetshort. Notably, we observe how in both the considered settings \methodshort{} outperforms the privileged approach relying on human-knowledge pre-defined attributes, showcasing how the \vlm effectively relies on its knowledge to predict a discriminative attribute fingerprint.

Tab.~\ref{tab:verification-myvlm} and Tab.~\ref{tab:verification-yollava} report the analysis on different verification strategies, following the experimentation reported in the main paper. MyVLM~\citep{alaluf2024myvlm} and \yollava~\citep{nguyen2024yollava} datasets are considered, respectively. Results show that ~\methodshort{} multimodal verification step outperforms the other compared strategies.

Finally, in Tab.~\ref{tab:retrieval-myvlm} (MyVLM~\citep{alaluf2024myvlm}) and Tab.~\ref{tab:retrieval-yollava} (\yollava~\citep{nguyen2024yollava}) consistent observations on the effectiveness of the proposed multimodal concept retrieval are reported.
\begin{table}
\centering
\label{table:ablation_inputs}
\resizebox{0.95\linewidth}{!}{  
\begin{tabular}{ccc ccc}
\toprule  
\makecell{\textbf{Pairwise} \\\textbf{Reasoning}} & \makecell{\textbf{Fingerprint}\\\textbf{Attributes}} & \makecell{\textbf{Reasoning}\\\textbf{CoT}} & \makecell{\textbf{Recognition}\\\textbf{Wtd}} & \makecell{\textbf{Captioning} \\ \textbf{Recall}} \\ 
\midrule

\textcolor{red}{\normalsize \ding{55}} & \textcolor{red}{\normalsize \ding{55}} & \textcolor{red}{\normalsize \ding{55}} & 96.1 & 87.4\\

\textcolor{red}{\normalsize \ding{55}} & \textcolor{red}{\normalsize \ding{55}} & \textcolor{Green}{\normalsize \ding{51}} & 94.2 & 88.8\\

\textcolor{Green}{\normalsize \ding{51}} & \textcolor{red}{\normalsize \ding{55}} & \textcolor{red}{\normalsize \ding{55}} & \textbf{98.7}  & 88.2\\

\textcolor{Green}{\normalsize \ding{51}} & \textcolor{red}{\normalsize \ding{55}}  &  \textcolor{Green}{\normalsize \ding{51}} & \underline{97.6} & \underline{89.6}\\

\cellcolor{method-green}\textcolor{Green}{\normalsize \ding{51}} & \cellcolor{method-green}\textcolor{Green}{\normalsize \ding{51}}  & \cellcolor{method-green}\textcolor{Green}{\normalsize \ding{51}} & \cellcolor{method-green}97.4 & \cellcolor{method-green}\textbf{91.5} \\

\cellcolor{tab-red}\textcolor{Green}{\normalsize \ding{51}} & \cellcolor{tab-red}privileged  & \cellcolor{tab-red}\textcolor{Green}{\normalsize \ding{51}} & \cellcolor{tab-red}97.8 & \cellcolor{tab-red}91.1 \\

\bottomrule  
\end{tabular}
}

\caption{Ablation on pairwise reasoning, fingerprint attributes, and the use of CoT reasoning for \inlineColorbox{method-green}{\methodshort{}} in terms of weighted recognition metrics ($\uparrow$) and captioning recall ($\uparrow$) on MyVLM~\cite{alaluf2024myvlm}. We include a \inlineColorbox{tab-red}{privileged} version of our approach with human pre-defined fingerprint attributes for concepts.}
\label{tab:ablation_myvlm}
\end{table}

\begin{table}
\centering
\label{table:grounding}
 \resizebox{0.62\linewidth}{!}{  
\begin{tabular}{lc}
\toprule  
\textbf{Method} & \textbf{Captioning Recall} \\
\midrule
Pairwise-reasoning & 91.3 \\
\hdashline
No estimation & 90.5 \\
Abstention & 90.4 \\
Logits-based & 90.8 \\
\cellcolor{method-green}Attr. Verification \textbf{(Ours)} & \cellcolor{method-green} \textbf{91.5} \\
\bottomrule  
\end{tabular}
}
\caption{Ablation on verification strategies for \inlineColorbox{method-green}{\methodshort{}} on MyVLM~\citep{alaluf2024myvlm} dataset. Performance is evaluated based on captioning recall ($\uparrow$).}
\label{tab:verification-myvlm}
\end{table}

\begin{table}
\centering
\label{table:ablation_inputs}
\resizebox{0.95\linewidth}{!}{  
\begin{tabular}{ccc ccc}
\toprule  
\makecell{\textbf{Pairwise} \\\textbf{Reasoning}} & \makecell{\textbf{Fingerprint}\\\textbf{Attributes}} & \makecell{\textbf{Reasoning}\\\textbf{CoT}} & \makecell{\textbf{Recognition}\\\textbf{Wtd}} & \makecell{\textbf{Captioning} \\ \textbf{Recall}} \\ 
\midrule  
\textcolor{red}{\normalsize \ding{55}} & \textcolor{red}{\normalsize \ding{55}} & \textcolor{red}{\normalsize \ding{55}} & 92.4 & 73.8 \\

\textcolor{red}{\normalsize \ding{55}} & \textcolor{red}{\normalsize \ding{55}} & \textcolor{Green}{\normalsize \ding{51}} & 89.1 & 76.8 \\

\textcolor{Green}{\normalsize \ding{51}} & \textcolor{red}{\normalsize \ding{55}} & \textcolor{red}{\normalsize \ding{55}} & 93.5  & \underline{83.4} \\

\textcolor{Green}{\normalsize \ding{51}} & \textcolor{red}{\normalsize \ding{55}}  &  \textcolor{Green}{\normalsize \ding{51}} & \textbf{95.4} & 81.5 \\

\cellcolor{method-green}\textcolor{Green}{\normalsize \ding{51}} & \cellcolor{method-green}\textcolor{Green}{\normalsize \ding{51}}  & \cellcolor{method-green}\textcolor{Green}{\normalsize \ding{51}} & \cellcolor{method-green}\underline{94.4} & \cellcolor{method-green}\textbf{87.1} \\ 

\cellcolor{tab-red}\textcolor{Green}{\normalsize \ding{51}} & \cellcolor{tab-red}privileged  & \cellcolor{tab-red}\textcolor{Green}{\normalsize \ding{51}} & \cellcolor{tab-red}96.7 & \cellcolor{tab-red}84.3 \\  
\bottomrule  
\end{tabular}
}

\caption{Ablation on pairwise reasoning, fingerprint attributes, and the use of CoT reasoning for \inlineColorbox{method-green}{\methodshort{}} in terms of weighted recognition metrics ($\uparrow$) and captioning recall ($\uparrow$) on \yollava~\cite{nguyen2024yollava}. We include a \inlineColorbox{tab-red}{privileged} version of our approach with human pre-defined fingerprint attributes for concepts.}
\label{tab:ablation-yollava}
\end{table}

\begin{table}
\centering
\label{table:grounding}
 \resizebox{0.62\linewidth}{!}{  
\begin{tabular}{lcc}
\toprule  
\textbf{Method} & \textbf{Captioning Recall} \\
\midrule
Pairwise-reasoning & 88.1 \\
\hdashline
No estimation & 82.0\\
Abstention & 85.6 \\
Logits-based & 83.4 \\
\cellcolor{method-green}Attr. Verification \textbf{(Ours)} & \cellcolor{method-green} \textbf{91.5} \\
\bottomrule  
\end{tabular}
}
\caption{Ablation on verification strategies for \inlineColorbox{method-green}{\methodshort{}} on Yollava~\citep{nguyen2024yollava} dataset. Performance is evaluated based on captioning recall ($\uparrow$).}
\label{tab:verification-yollava}.
\end{table}

\begin{table}[t]
\centering

\resizebox{0.8\columnwidth}{!}{%
\begin{tabular}{lcccc}
\toprule
\textbf{Embedding} & \textbf{H@1} & \textbf{H@3} & \textbf{H@5} & \textbf{H@10} \\
\midrule
DINOv2 & 64.7 & 80.3 & 86.5 & 93.8 \\
CLIP-Image & 88.2 & 96.2 & 98.2 & \underline{99.4} \\
CLIP-Text & 80.5 & 95.6 & 98.1 & 98.7 \\
Multi-modal (2-step) & \underline{92.2} & \underline{98.2} & \underline{98.8} & \underline{99.4} \\
\cellcolor{method-green}\methodshort{} \textbf{(Ours)} & \textbf{93.5
} & \textbf{98.5} & \textbf{99.4} & \textbf{100} \\
\bottomrule
\end{tabular}%
}
\caption{Performance with different retrieval strategies evaluated in terms of HIT@K ($\uparrow$) on MyVLM~\citep{alaluf2024myvlm} dataset.}
\label{tab:retrieval-myvlm}
\end{table}

\begin{table}[]
\centering
\resizebox{0.8\columnwidth}{!}{%
\begin{tabular}{lcccc}
\toprule
\textbf{Embedding} & \textbf{H@1} & \textbf{H@3} & \textbf{H@5} & \textbf{H@10} \\
\midrule
DINOv2 & 67.3 & 83.5 & 88.3 & 94.3 \\
CLIP-Image & 82.9 & 93.4 & 94.9 & 100 \\
CLIP-Text & 64.8 & 85.8 & 94.7 & \underline{99.4} \\
Multi-modal (2-step) & \textbf{92.2} & \underline{98.2} & \underline{98.8} & \underline{99.4} \\
\cellcolor{method-green}\methodshort{} \textbf{(Ours)} & \underline{84.7
} & \textbf{99} & \textbf{99.6} & \textbf{100} \\
\bottomrule
\end{tabular}%
}
\caption{Performance with different retrieval strategies evaluated in terms of HIT@K ($\uparrow$) on \yollava~\citep{nguyen2024yollava} dataset.}
\label{tab:retrieval-yollava}
\end{table}

\end{document}